\documentclass[letterpaper]{article} 
\usepackage{aaai25}  
\usepackage{times}  
\usepackage{helvet}  
\usepackage{courier}  
\usepackage[hyphens]{url}  
\usepackage{graphicx} 
\usepackage{natbib}  
\usepackage{caption} 
\usepackage{algorithm}
\usepackage{url}
\usepackage{amsmath}
\usepackage{amssymb}
\usepackage{booktabs}
\usepackage{epsfig}
\usepackage{color}
\usepackage{xcolor}
\usepackage{comment}
\usepackage{tikz}
\usepackage{multirow}
\usepackage{svg}
\usepackage{subcaption}
\usepackage{multirow}
\usepackage{array} %
\usepackage{rotating}
\usepackage{soul}
\usepackage{times,xspace} 
\usepackage{array}
\usepackage{enumitem}
\usepackage{xr}
\usepackage{subcaption}
\usepackage{algorithm}
\usepackage{algpseudocode}
\usepackage{blindtext}
\usepackage{makecell}
\usepackage[T1]{fontenc} 
\usepackage[utf8]{inputenc}
\newcommand{\jaisidh}[1]{\textcolor{blue}{[jaisidh:~#1]}}
\newcommand{\harsh}[1]{\textcolor{purple}{[harsh:~#1]}}

\usepackage{newfloat}
\usepackage{listings}
\DeclareCaptionStyle{ruled}{labelfont=normalfont,labelsep=colon,strut=off} 
\floatstyle{ruled}
\newfloat{listing}{tb}{lst}{}
\floatname{listing}{Listing}
\title{Automatic Discovery and Assessment of Interpretable Systematic Errors in Semantic Segmentation}
\author{
    Jaisidh Singh\equalcontrib\thanks{Work done during an internship at Bosch.}\textsuperscript{\rm 1},
    Sonam Singh\equalcontrib\textsuperscript{\rm 2},
    Amit Arvind Kale\textsuperscript{\rm 2},
    Harsh K Gandhi\textsuperscript{\rm 3}
}
\affiliations{
    \textsuperscript{\rm 1}University of Tübingen,
    \textsuperscript{\rm 2}Robert Bosch Corporate Research India,
    \textsuperscript{\rm 3}California Institute of Technology\\
    \vspace{1pt}
    {\tt\small{jaisidh.singh@student.uni-tuebingen.de}},
    {\tt\small{\{sonam.singh,amitarvind.kale\}@in.bosch.com}},
    {\tt\small{hgandhi@caltech.edu}}
}

\usepackage{bibentry}

\begin{document}

\maketitle

\begin{abstract}
    This paper presents a novel method for discovering systematic errors in segmentation models. For instance, a systematic error in the segmentation model can be a sufficiently large number of misclassifications from the model as a parking meter for a target class of pedestrians. With the rapid deployment of these models in critical applications such as autonomous driving, it is vital to detect and interpret these systematic errors. However, the key challenge is automatically discovering such failures on unlabelled data and forming interpretable semantic sub-groups for intervention. For this, we leverage multimodal foundation models to retrieve errors and use conceptual linkage along with erroneous nature to study the systematic nature of these errors. We demonstrate that such errors are present in SOTA segmentation models (UperNet ConvNeXt and UperNet Swin) trained on the Berkeley Deep Drive and benchmark the approach qualitatively and quantitatively, showing its effectiveness by discovering coherent systematic errors for these models. Our work opens up the avenue to model analysis and intervention that have so far been underexplored in semantic segmentation.
\end{abstract}
%
%
\section{Introduction}
\label{introduction}
Advances in deep learning have led to rapid growth in the performance of computer vision models. In particular, Autonomous Driving (AD) as an application has attracted much attention in the context of using computer vision \cite{feng2020deep}. State-of-the-art AD vehicles use segmented maps of the natural world to "see" like a human. 
These segmented maps are obtained by using Semantic Segmentation Models (SSMs), which aim to ensure the semantic understanding of every pixel in the input image taken from the camera~\cite{wang2018understanding, yu2018bisenet}.
A required standard for AD vehicles is that the SSMs must be able to perceive the visual input with extremely high precision and accuracy. This is particularly important considering human safety is of utmost priority while designing the models for AD vehicles. 
%
%
To ensure this, recent research in this field, as evidenced by \cite{acdc}, has focused on developing and implementing robust SSMs capable of satisfactorily meeting benchmark metrics for accurate predictability \cite{xie2017adversarial}. 
%
%

However, there is a need to go beyond benchmark metrics in order to understand the weaknesses of SSMs. Specifically, knowing where an SSM fails can reveal internal biases of the model as well as spurious correlations learnt during training. Hence, the analysis of \emph{systematic errors} made by SSMs emerges as an important line of investigation.
%
%
Systematic errors are groups of hard samples, or \emph{slices} of data, on which a model would show significantly lower performance as compared to other samples. 
%
%
Recently, Slice Discovery Methods (SDM) have shown potential to pinpoint systematic subsets by identifying coherent sub-groups of errors through analysis of the performance of the model across all samples of the test dataset. However, like most of the literature on systematic error assessment, the SDMs primarily focus on image classification models \cite{eyuboglu2022domino, jain2022distilling, sohoni2020no}, leaving the analysis for SSMs largely unexplored. Further, SSMs face unique challenges in the analysis of systematic errors such as:

\begin{enumerate}
    \item \textbf{Scarcity of labelled test data}: Studying systematic errors based on metrics given by ground truth of the test data becomes impractical for SSMs, due to the dearth of clean, labelled test data in real-world AD scenarios.
    %
    %
    \item \textbf{Random vs. human-interpretable:} It is easy to understand errors in image classifiers as there is only one semantic considered per image. However for SSMs, an image may contain several regions which do not denote anything (random blurs, black regions etc.). These seemingly \textit{random errors} may be systematic, however, downstream human analysis would likely not reveal impactful reasons for why the SSM failed there. Hence, finding human-interpretable systematic errors will be critical in the downstream rectification of the SSM.
\end{enumerate}

This work aims to fill the gaps mentioned above, and presents an automated framework which discovers and explains interpretable systematic errors in AD settings. 
This is made possible by leverage foundational models to interpret SSM predictions at the concept level. Hence, our purely inference-driven framework can be applied to large-scale, unstructured and unlabelled data domains. Overall, our study focuses on three structural foundations:
\begin{enumerate}
    \item \textbf{Concept-level Interpretability:} Our method analyses the outputs of SSMs at conceptual levels by using foundation models to study linked content within errors. This removes the need for tedious pixel-wise labelling.
    \item \textbf{Zero-shot nature:} The proposed framework alleviates the cumbersome human-based evaluation of SSMs to discover systematic errors in a training-free, purely inference-based manner, even in the case of largely, unstructured and unlabelled data. 
    \item \textbf{Compatibility with any SSMs:} The proposed framework allows for universal adaption of the systematic error detection and rectification for any SSMs. Further, the foundation models used can also be changed, making the framework fully modular.
\end{enumerate}
As a proof of concept, we evaluate our framework on state-of-the-art SSMs trained on Berkeley Deep Drive~\cite{yu2020bdd100k} and benchmark the approach both qualitatively and quantitatively.
Our results demonstrate the effectiveness of the proposed framework in discovering and interpreting systematic errors specific to the given SSMs.
To the best of our knowledge, our work is the first to propose a solution applicable at the broad scale of any SSM and unstructured data.
Additionally from a safety engineering perspective, the human-interpretable assessments provided by our framework can allow it to be utilised as a necessary step for evaluating ready-to-deploy semantic segmentation models.
%
%
\section{Background}
\subsection{Systematic error detection}
Several studies~\cite{sohoni2020no, d2022spotlight} have worked towards using model representations for discovering continuous regions of poor performance without the available sub-group labels. Furthermore, a recent model, DOMINO \cite{eyuboglu2022domino}, finds the underperforming slices using mixture models and exploits the multimodal embedding space of CLIP~\cite{clip} to match the most relevant caption to describe the found slices. Jain et al. \cite{jain2022distilling} use linear classifiers (e.g., SVM \cite{gandhisvm}) to model the failure modes as directions in a visual-language latent space and leverage text-to-image generative models for synthetic targeted data augmentation. 
Furthermore, multiple studies ~\cite{xia2020synthesize, guillory2022erroraug, marufur2021fsnet} have investigated errors in SSMs which maybe due to labelling, etc. However, these do not align with our work as they lack the study of the systematic nature of these errors and their interpretability, and do not operate zero-shot.
\subsection{Foundational Models}
Recently, foundation models have emerged as highly adaptable modules for various downstream tasks due to their pretraining on large-scale datasets and the versatility of fine-tuning. Their inception can be traced back to Large Language Models (LLMs) like BERT~\cite{devlin2018bert}, GPT-2~\cite{radford2019language}, T5~\cite{raffel2020exploring}, GPT-3~\cite{brown2020language} etc. Since language is a rich abstraction of the real world, these models exhibit a remarkable understanding of real-world concepts. Additionally, the utility of the transformer architecture brought forth seminal multimodal works like VisualBERT~\cite{li2019visualbert}, CLIP, Florence~\cite{yuan2021florence}, BLIP~\cite{li2022blip} emerged as visual-language models. These recognize visual and language concepts for application in image-text retrieval, visual question answering (VQA), natural language visual reasoning (NLVR), etc. Subsequent works utilize this multimodal paradigm in the localization or detection regime, exemplified by models like GLIP~\cite{li2022glip} and GroundingDINO~\cite{groundingdino}. Our paper uses cross-domain transfer abilities of multimodal foundation models to interpret systematic errors at the concept level. 
\label{related_work}

%
\begin{figure*}[!t]
	 {\includegraphics[width=\textwidth] {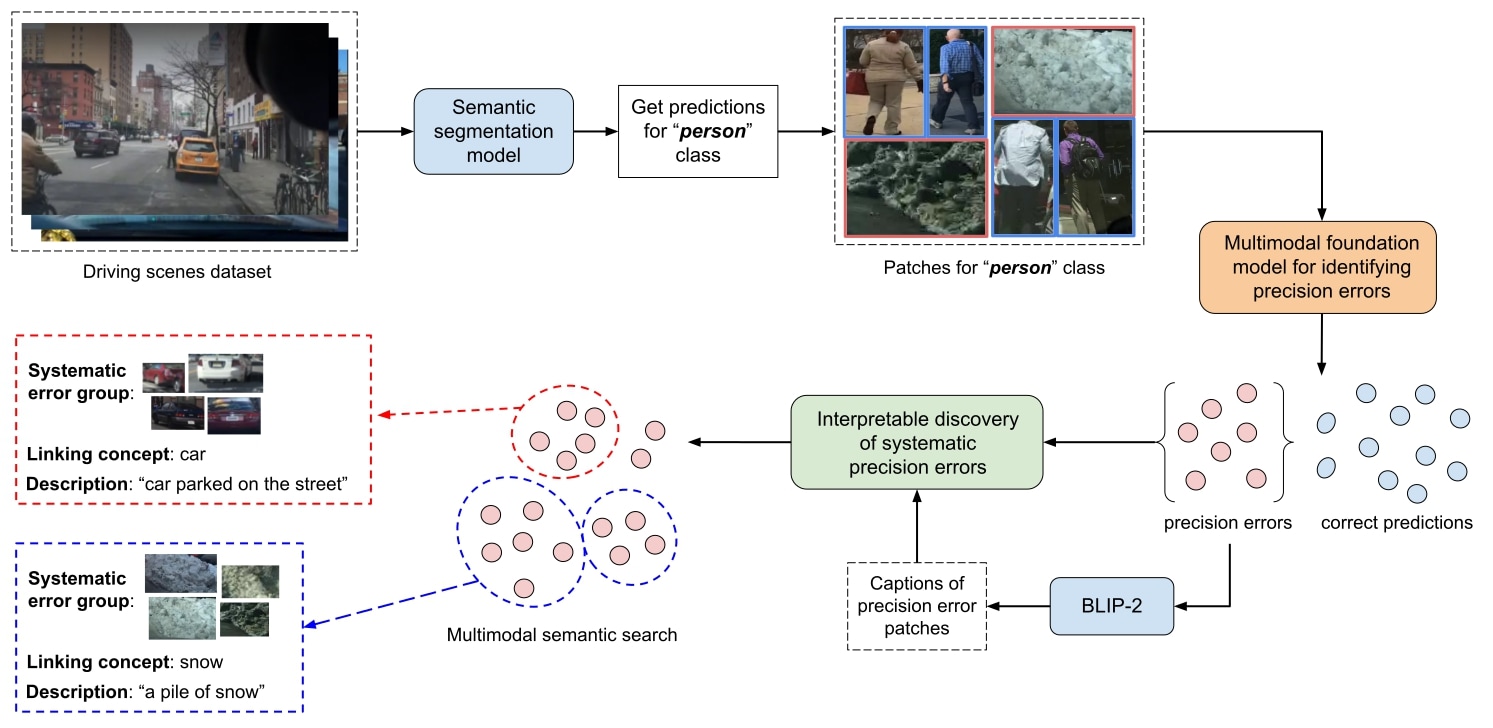}}
	\caption{Our framework begins with the inference of an SSM w.r.t. a particular semantic class $c_j$, ``\textit{person}'' in this case. Image regions corresponding to the dense predictions for $c_j$ are extracted as \textit{patches} and are fed to a multimodal foundation model. This model identifies patches which do not represent $c_j$, \textit{i.e.}, patches which denote precision errors. Finally, these precision errors are utilized by an algorithm in order to reveal systematic error groups denoting a common human-interpretable concept.}
	\label{fig:overview_large}
\end{figure*}
\section{Methodology}
We present our approach in three stages. First, we outline the assumptions and settings that our framework operates on. Next, we describe our procedure for detecting errors made by the SSM. Finally, we present our algorithm to identify which errors are systematic as well as human-interpretable. A visual overview of our method is given in Fig.~\ref{fig:overview_large}.
%
%
\subsection{Preliminaries}
\label{sec:prelims}
Before going into the details of our method, let us first understand the errors that an SSM can make. An SSM, denoted by $f(\cdot)$, operates on an image, or a group of pixels $X$, and assigns a semantic class $c_i$ to each pixel. Here $c_i$ is one of $m$ possible classes, \textit{i.e.} $i \in \{1, ..., m\}$.
Taking a driving scene for instance, where the desired semantic class $c_i$ is ``\textit{person}'', if the SSM prediction $f(X)$, is erroneous, it can be called a:
%
%
\begin{enumerate}
    \item \textbf{Recall error (false negative prediction)}: If $f(X)$ is misclassified as $c_j$ for some $j \ne i$, when the pixels in reality belong to $c_i$, hence resulting in a false negative prediction. For example, say the segmentation model misclassified a {person} in the image as a ``\textit{fire hydrant}''.

    \item \textbf{Precision error (false positive prediction)}: If $f(X)$ is misclassified as $c_i$, when the pixels actually belong to $c_k$, for some $i\ne k$, hence resulting in a false positive prediction. In the case of our example, say the segmentation model misclassifies a {traffic pole} to be a ``\textit{person}''.
\end{enumerate}
%
%
This paper limits itself to studying systematic nature of \emph{precision errors} for vulnerable road users (VRUs), specifically the ``\textit{person}'' and ``\textit{bicycle}'' classes in driving scenes. 
As mentioned earlier, there can be numerous random regions which can influence an SSMs decision and which can act as systematic errors. However, they add little value in downstream intervention. Hence, we restrict our assessment to human-interpretable systematic errors for which downstream analysis can be coherently performed.
\subsection{Retrieval of Precision Errors in SSM Predictions}
\label{sec:finderror}
In the entire framework, we work with the assumption that we don't have access to labelled data. Identifying precision errors in the latter case would be rather trivial. In our case however, the only data we have to work with are the raw RGB image dataset. To bypass this issue, we retrieve precision errors for a particular semantic class $c_j$ for $j \in \{1, ..., m \}$, using image \textit{patches} that we get from using the procedure elucidated below. 
%
%
\subsubsection{Defining and obtaining Patches}
\label{patches}
Formally, a patch within an image $I_i$, denoted by $p_i^j$, encompasses the image content inside the bounding box of a set of neighboring pixels $X_{ik}^j$ classified by the SSM $f(X_{ik}^j)$ as class $c_j$. The subscript $i$ in $p_i^j$ indicates the originating image $I_i$, while the superscript $j$ specifies the semantic class assigned by the SSM. Notably, the subscript $k$ in $X_{ik}^j$ signifies the total number of patches in image $I_i$, a figure we assume remains constant across all images.
Now within each predicted class map $\hat{Y}_i$, numerous regions of pixels may be uniformly classified as belonging to semantic class $c_j$ by $f(\cdot)$. To extract these patches, we employ the following methodology:
\begin{enumerate}
    \item \textbf{Step 1:} For each region classified as semantic class $c_j$ in the RGB image $I_i$, we isolate the image content within the bounding box. This procedure yields the patches $p^j_i$s that the SSM classifies as $c_j$.
    \item \textbf{Step 2:} Step 1 is repeated for all images in the dataset to compile a set $\boldsymbol{P}^j = \{p_i^j\}_{i=0}^{N}$, encapsulating all patches identified as class $c_j$ by the SSM across $N$ images.
\end{enumerate}
It is crucial to recognize that the patches in $\boldsymbol{P}^j$ may contain image content accurately or inaccurately classified as $c_j$ by the SSM. Therefore, our subsequent objective is to delineate the SSM's correct predictions from its errors, subsequently isolating the precision errors therein.
%
%
%
\subsubsection{Precision Error Identification as Binary Classification}
%
%
\label{Grounding_dino}

The SSM-predicted patch would constitute as a precision error if the said patch belongs to the set $\boldsymbol{P}^j$ but does not depict an object of semantic class $c_j$ at the concept level.
This decision rule turns precision error identification into a binary classification problem, where samples belonging to the \textit{positive class} represent the concept of $c_j$, while those belonging to the \textit{negative class} do not. 
To note, this binary classification scheme must take into account that the framework must work on unlabelled data and must also be completely training-free.
%
%
%
Hence, we leverage a foundation model as the binary classifier, denoted as $g(\cdot, \cdot)$.
This model is in fact a multimodal object detector, and effectively integrates the localization of concepts in images with their semantic labels in natural language. This property caters to the binary classification task in hand. Further, the extensive pre-training of this model affords a comprehensive understanding of real-world concepts and can thus be used without any training. Consequently, it can viably detect the presence of the semantic class $c_j$ within a given patch. 
%
%
%

For the classification routine, every patch in the set $\boldsymbol{P}^j$ is fed to the model as input and gets the output given by $\{B_l\} = \{g(p^j_i, c_j)\}_{j=0}^l$ for all $l$ patches in the $\boldsymbol{P}^j$  which correspond to the class $c_j$. Here, $\{B_l\}$ is the set of bounding boxes that localize the concept of $c_j$ within the patch.
%
%
Our criterion for identifying a precision error within a patch is to examine whether any bounding box was predicted by $g(., .)$. Specifically, if no bounding boxes are predicted ($B_l = \phi$, where $\phi$ denotes an empty set), the patch $l$ is considered to belong to the negative class.
Conversely, if a patch has bounding box predicted using $g(., .)$, it denotes the presence of $c_j$ and is therefore not an error (belongs to the \textit{positive class}). Hence, we obtain a set $\boldsymbol{\tilde{P}}^j = \{ 
p_i^j  \enspace | \enspace g(P_i^j, c_j) = \phi, \enspace p_i^j \in \boldsymbol{P}^j \}$ for all $i$ constituting the patches identified as precision errors by $g(\cdot, \cdot)$. Now with the precision error patches, we can proceed to the final stage of our framework: assessing systematicity in the errors. 

%
\subsection{Interpretable Systematic Error Discovery}
\label{sec:mining}
%
%
Identifying systematic nature of the errors $\boldsymbol{\tilde{P}}^j$ is the next part of the framework. For this, we construct an algorithm using both image and language modalities to interpretably predict the systematic nature of the precision error. This is because language allows us to describe the predicted systematic error groups. Simultaneously, it allows us to verify the concepts shared among the systematic error samples in a rich representation space.
%
%
\subsubsection{Evaluation procedure of the obtained precision errors}
\label{clip_embeddings}
The first step is to evaluate the obtained precision errors from the classification routine explained in the previous subsection. For this, we examine each \textit{query patch} $p^j_i$ in $\boldsymbol{\tilde{P}}^j$ and determine its $q$-nearest neighbors within $\boldsymbol{\tilde{P}}^j$, denoted by $\mathcal{N}$. 
This process occurs within the latent space of CLIP, where the CLIP image embedding of patch $p_i^j$ is given by $E_p = h_{img}(p)$. Here, $h_{img}(\cdot)$ denotes the CLIP image encoder. Furthermore, the CLIP image embeddings of the patches in $\mathcal{N}$ are encompassed in the set $\boldsymbol{E}_\mathcal{N}$ given by $\boldsymbol{E}_{\mathcal{N}}= \{h_{img}(v) \text{ } | \text{ } v \in \mathcal{N} \}$. 
Next, a pre-trained BLIP-2~\cite{li2023blip} model generates captions of the \textit{query patch} and its neighboring patches.
These captions are represented by $T_p$ for the query patch and $\boldsymbol{T}_\mathcal{N}$ for the neighboring patches. Finally, for a given query patch to be categorized as an interpretable systematic error, it must

\begin{enumerate}
    \item  \textbf{Criterion 1}: Show strong \textbf{conceptual linkage} with nearest neighbouring patches. Essentially, when a patch and its neighboring patches collectively represent a highly similar or interconnected concept, and yet the model misclassifies the given patch, it demonstrates the systematic tendency of the model to misclassify the concept itself. Moreover, establishing conceptual connections with neighboring patches helps distinguish systematic errors from random ones, making them more interpretable. In this study, interpretablility necessitates a criterion for our decision-making process as described above.
    \item \textbf{Criterion 2}: Must be consistently an \textbf{error }\textit{i.e.}, it must not represent $c_j$. Consistent misclassification of patches relating to the semantic class $c_j$ represent a systematic issue w.r.t the model in predicting the concept.
\end{enumerate}

Establishing conceptual linkage can be complicated and therefore, will require an added layer of complexity. Considering we want to keep the model in a training free-paradigm, we devise a series of procedures aimed at establishing conceptual linkages with the nearest neighbors, as elaborated in Section 3.3.2. Subsequently, in Section 3.3.3, we present a mechanism for quantifying the criterion 2. Finally, in Section 3.4, we integrate these criteria to make a final determination regarding the systematicity of an error.
%
%
\subsubsection{Conceptual linkage with nearest neighbors}
\label{mpnet}
We establish conceptual linkage in both the patches' image space and the generated captions' language space. Although having conceptual linkage solely in the image space ensures consistency of patch concepts among neighbors, it does not verify the consistency of the generated explanations. 
%
%
Hence, we include both modalities in the systematicity criteria.
%
%
\paragraph{Image space of patches:} For the image space conceptual linkage, we compute the CLIP text embedding of $T_p$, given by $E'_p = h_{text}(T_p)$ where $h_{text}(\cdot)$ is the CLIP text encoder. Next, the conceptual similarity measure $\sigma_1$ is computed as
\begin{equation}
    \sigma_1 = \frac{1}{q} \sum_{E \in \boldsymbol{E}_{\mathcal{N}}} s(E'_p, E)
    \label{eq:s1}
\end{equation}
where $s(\cdot, \cdot)$ denotes the cosine similarity function. Higher values of $\sigma_1$ denote strong conceptual similarity of the nearest neighbor patches to the content of the query patch.
%
\paragraph{Language space of captions:}  To evaluate conceptual linkage in the language space, we utilize a sentence encoder $\Phi(\cdot)$, which is a pre-trained language model used for embedding text for semantic search. 
Particularly, $\Phi(\cdot)$ is used to combine $T_p$ and $\boldsymbol{T}_\mathcal{N}$ into the embeddings $\mathcal{E}_p = \Phi(T_p)$ and $\boldsymbol{\mathcal{E}}_{\mathcal{N}} = \{\Phi(t) \text{ } | \text{ } t \in \boldsymbol{T}_\mathcal{N} \}$ respectively.
Similar to above, we compute $\sigma_2$ by
\begin{equation}
    \sigma_2 = \frac{1}{q}\sum_{\mathcal{E} \in \boldsymbol{\mathcal{E}}_{\mathcal{N}}} s(\mathcal{E}_p, \mathcal{E})
    \label{eq:s2}
\end{equation}
Likewise, $\sigma_2$ signifies the semantic similarity between the language descriptions of the query patch and those of its nearest neighbor patches.
Using high $\sigma_1$ and $\sigma_2$ measures, one can demonstrate high conceptual linkage of the query patch to its nearest neighbors, thereby satisfying criterion 1. 

\subsubsection{Erroneous nature of query patch}
To satisfy criterion 2, a text prompt is constructed for $c_j$, given by $t_j = $ ``\textit{the concept of one or many} \texttt{\{$c_j$\}}''. This is subsequently converted into the embedding $\mathcal{E}_j = \Phi(t_j)$ and is used with $\mathcal{E}_p$ to compute $\sigma_3$ using
\begin{equation}
    \sigma_3 = s(\mathcal{E}_p, \mathcal{E}_j).
\end{equation}
For the query patch to be a systematic error, it must not represent the concept of $c_j$; hence $\sigma_3$ must be low in value. Despite the fact that we work to evaluate systematic errors in a set of patches which have been classified as errors in the previous section, we implement this to filter out any mistakes made while identifying precision errors. 
%
%
This provides a validation step to the classification before, and also fine-tunes the results of our overall framework.
%
%
\subsection{Final determination of systematic errors}
\label{final_determination}
Applying the defined conditions and utilizing the derived values of $\sigma_1$, $\sigma_2$, and $\sigma_3$ metrics, we can ultimately infer the systematic characteristics of the queried patch $p$ through the function $\Omega(\cdot)$, as outlined by:
%
%
\begin{equation}
    \Omega(p) = 
    \begin{cases}
    1 & \text{if } \sigma_1 + \sigma_2 - \sigma_3 \geq \alpha\\
    
    0 & \text{otherwise }.
\end{cases}
\end{equation}
Here, $\alpha$ is a scalar threshold for the value of $\sigma_1 + \sigma_2 - \sigma_3$. This threshold is determined empirically and is discussed in the subsequent section. Nonetheless, the output to the testing $\Omega(p) = 1$ denotes that the query patch is a \emph{interpretable} systematic error, dismissing the possibility of the error being arbitrary. In contrast, $\Omega(p) = 0$ indicates that the patch denotes a random (un-interpretable) systematic error, or an error which is simply not systematic.
%

%
%

        




%
\section{Experiments}
%
%
%
%
We organise our experimental approach in three sections. The first section is dedicated to details of the datasets we use and outlines the settings for semantic segmentation in our experiments. Next, we experiment with identifying precision errors within the datasets using $g(\cdot, \cdot)$. Finally, the final section presents a comprehensive analysis of systematic error assessment, identification, and subsequent explanations. 
%
%
\label{experiments}
\subsection{Datasets and SSM Settings}
\label{sec:semseg_settings}
\paragraph{Semantic segmentation datasets:} We conduct experiments using two significant real-world benchmarks datasets for semantic segmentation:
%
    \textit{(i) Berkeley DeepDrive Dataset (BDD) }~\cite{yu2020bdd100k}: The BDD dataset offers a large collection of images taken from vehicle-mounted cameras in urban and rural locations across North America. In our work, we utilize the \emph{labelled subset} of the dataset tailored for semantic segmentation 
    %
    %
    and so, our setting has 7000 labelled training images with 1000 labelled validation images amalgamated into a unified set. This combined set of 8000 images are annotated and are classified into one of the 19 distinct semantic classes. 
    \textit{(ii) Adverse Conditions Dataset (ACDC)}~\cite{acdc}:
    The {A}dverse {C}onditions {D}ataset with {C}orrespondences features 4,006 images of driving scenarios under challenging visual conditions, such as fog, snow, rain, and nighttime. It also provides equivalent scenes in daylight and clear weather, with the same pixel-level annotation as the BDD dataset.
%
\paragraph{Model and Test Settings:} We conduct experiments using two different SSMs namely the \textit{ConvNeXt}~\cite{convnext} model and the \textit{Swin Transformer}~\cite{liu2021swin}. 
%
%
For our experiments, the SSMs were trained on the BDD dataset using the UperNet~\cite{upernet} framework and a training routine similar to that described in Wang et al. (\cite{wang2023upercx}). Specifically, these were obtained in a pretrained manner~\cite{bddrepo}. 
We choose these SSMs and datasets to evaluate our framework in two settings per SSM: \textit{in-distribution} (or in-sample) and \textit{out-distribution} (or out-sample). The in-distribution test data consists of BDD images on which the models were trained. For out-distribution test, we used the ACDC dataset. This practice aims to provide insight into the types of systematic errors afforded by seen and unseen distributions.
%
%
%
\subsection{Precision Error Identification}
\label{sec:detector_benchmark}
%
%
In this section, we study how $g(\cdot, \cdot)$ can predict precision errors from patches. 
We broadly benchmark the ability of $g(\cdot, \cdot)$ to correctly identify precision errors using an experimental setup which is presented as follows. 
%
%
\paragraph{Experimental Setup:} Following our methodology, $g(\cdot, \cdot)$ a multimodal object detection model, operates on each patch $p \in \boldsymbol{P}^j$ to predict a precision error if $g(p, c_j) = \phi$, indicating that no bounding boxes are identified for the semantic class $c_j$. Here, we select two choices for $g(\cdot, \cdot)$, GroundingDINO~\cite{groundingdino} and Owl-ViT~\cite{owlv2}. Both of these models are used in their pretrained state, via publically available checkpoints. When employing GroundingDINO as $g(\cdot, \cdot)$, we follow its pretrained configuration where the bounding box threshold is set to $0.35$ and the text prediction threshold is set to $0.25$. These thresholds denote the minimum required similarity to the semantic class name to initiate a prediction of a box and its label. Similarly when Owl-ViT is used, its minimum similarity threshold is set to $0.25$.
%

%
\paragraph{Evaluation:} We evaluate the precision errors identified by our framework by using the accuracy metric here (precision, recall, and f1-scores given in the appendix). Specifically, if the ground truth for a patch $p \in \boldsymbol{P}^j$ indicates a significant present of the semantic class $c_j$, yet $g(p, c_j) = \phi$, then a precision error is incorrectly detected (false positive). Conversely, if the ground truth indicates that the concept of $c_j$ is largely absent from $p$ and $g(p, c_j) \neq \phi$, then a precision error is incorrectly rejected (false negative). We quantify the extent of presence or absence of $c_j$ with the intersection over union (IOU) metric between $p$ and its ground truth. An algorithm for this evaluation is provided in the appendix.
%
%
%
%
\begin{table}[t]
    \centering
    \begin{tabular}{llcccc}
        \toprule
        \multirow{2}{*}{\small{Dataset}} & \multirow{2}{*}{$c_j$} & \multicolumn{2}{c}{\small{GroundingDINO}} & \multicolumn{2}{c}{\small{Owl ViT}}\\
        \cline{3-4} \cline{5-6}
         &  & \small{ConvNeXt} & \small{Swin} & \small{ConvNeXt} & \small{Swin}\\
         \midrule
        \multirow{2}{*}{\small{BDD}} & {\small{person}} & 69.77 & 66.85 & 69.66 & 64.83\\
         & {\small{bicycle}} & 65.44 & 70.28 & 72.35 & 75.1\\
        \multirow{2}{*}{\small{ACDC}} & {\small{person}} & 61.75 & 59.93 & 68.92 & 63.54\\
         & {\small{bicycle}} & 49.45 & 65.88 & 53.85 & 60.00\\
        \bottomrule
    \end{tabular}
    \caption{Accuracy of GroundingDINO and Owl-ViT in identifying precision errors where minimum patch size is $60 \times 60$.\vspace{-10pt}}
    \label{tab:detector_bench}
\end{table}
%
%
%
%
\paragraph{Results:} The outcomes of this study are detailed in Table.~\ref{tab:detector_bench}, which outlines the accuracy with which precision errors are identified in $\boldsymbol{P}^j$ for both the models and all datasets, specifically for the ``\textit{person}'' and ``\textit{bicycle}'' classes. 
GroundingDINO demonstrates somewhat lower efficacy on the ACDC dataset compared to other SSMs, possibly due to the challenging weather conditions in the ACDC images adversely affecting its recognition capabilities, especially in low-resolution patch settings. In contrast, across various models, data configurations, and input patch sizes for foundational models, Owl ViT consistently surpasses GroundingDINO in precision error detection according to accuracy measures. Notably, both detection models exhibit lower recall compared to precision, showing a tendency to classify patches indicating the concept of \(c_j\) as errors more frequently than actual errors. This observation underscores the rationale for criterion 2 above, emphasizing the analysis focus on true errors (do not denote the concept of \(c_j\)) only.
\begin{figure*}[t]
    \centering
    \includegraphics[width=\textwidth]{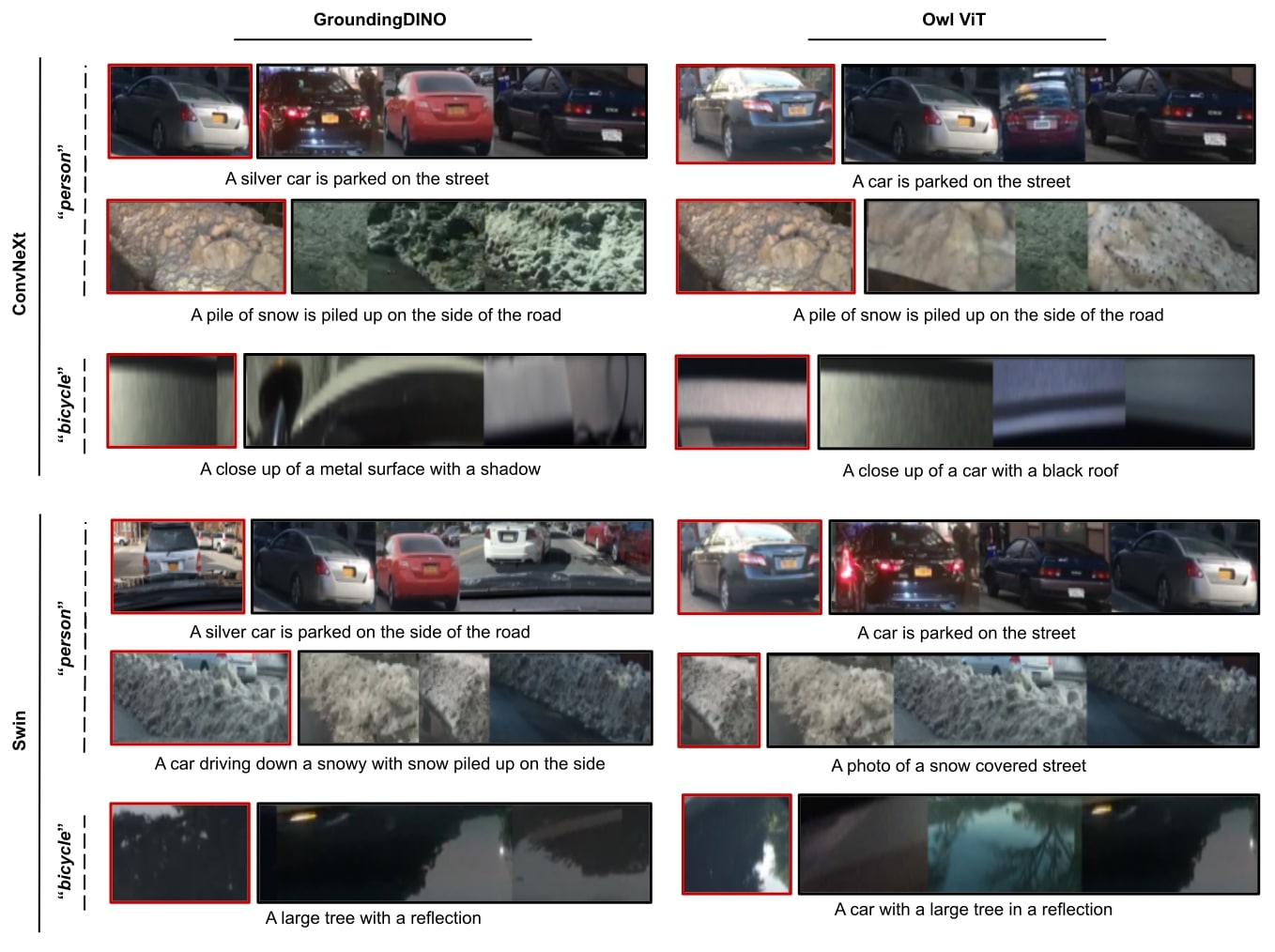}
    \caption{
    Qualitative assessment of systematic errors in the BDD dataset. 
    For the ``\textit{person}'' class, concepts of \textit{snow} and \textit{car} are systematically present in precision errors, while for the ``\textit{bicycle}'' class, the SSMs systematically err on \textit{car} and \textit{metal} parts.
    \vspace{-10pt}}
    \label{fig:bdd_person_qual}
\end{figure*}
\subsection{Finding Interpertable Systematic Errors}
We now present the final step of our framework: isolating interpretable systematic errors from $\boldsymbol{\tilde{P}^j}$, the set of patches predicted as precision errors by $g(\cdot, \cdot)$.
\paragraph{Experimental setup:} For each each query patch $p \in \boldsymbol{\tilde{P}}^j$, we predict whether it denotes an interpretable systematic error or not, \textit{i.e.}, if $\Omega(p) = 1$. The threshold $\alpha$ utilised in computing $\Omega(p)$ is set to $0.35$. This value is ascertained empirically: by observing the ranges of cosine similarities in each latent space used. This is necessary as different spaces can exhibit different cosine similarity ranges in order to match to small neighbourhoods of non-spurious concepts. Lastly, the size of the nearest-neighbourhood, $q$, is set to $3$ here.
\paragraph{Evaluation:} Our framework is designed to automatically discover human-interpretable systematic errors without using any ground truths, and so its evaluation requires a human intervention. To this end, we conduct a human study in which evaluators analyse the predictions of our framework. A patch, predicted to be an interpretable systematic error, is verified to be so only if the evaluator finds all of the following conditions to hold: (i) the query patch image depicts a concept different from $c_j$ and is coherent to human understanding, (ii) the nearest neighbour patches depict the same human-coherent concept, (iii) the caption for the query patch adequately represents the content of the query patch and its nearest neighbours. Overall, we use 3 evaluators such that an annotation is made only when 2 or more evaluators agree. Their assessments are thus aggregated to give the accuracy of finding interpretable systematic errors.
%
%
%
\begin{table}[h!]
    \centering
    \begin{tabular}{llcccc}
        \toprule

        \multirow{2}{*}{\small{Dataset}} & \multirow{2}{*}{$c_j$} & \multicolumn{2}{c}{\small{GroundingDINO}} & \multicolumn{2}{c}{\small{Owl ViT}}\\
        \cmidrule{3-4} \cmidrule{5-6}
         & & {\small{ConvNeXt}} & {\small{Swin}} & {\small{ConNeXt}} & {\small{Swin}}\\
         \midrule
        \multirow{2}{*}{\small{BDD}} & {\small{person}} & 95.23 & 89.55 & 84.86 & 82.67\\
         & {\small{bicycle}} & 80.72 & 69.00 & 74.48 & 67.84\\
        \multirow{2}{*}{\small{ACDC}} & {\small{person}} & 84.21 & 93.54 & 83.87 & 88.06\\
         & {\small{bicycle}} & 81.25 & 58.33 & 75.00 & 57.14\\
        \bottomrule
    \end{tabular}
    \caption{Accuracy in assessing interpretable systematic errors from the predictions of GroundingDINO and Owl-ViT, given minimum patch size as $60\times60$. \vspace{-10pt}}
    \label{tab:sys_acc}
\end{table}
\begin{figure*}[t]
    \centering
    {\includegraphics[width=\linewidth]{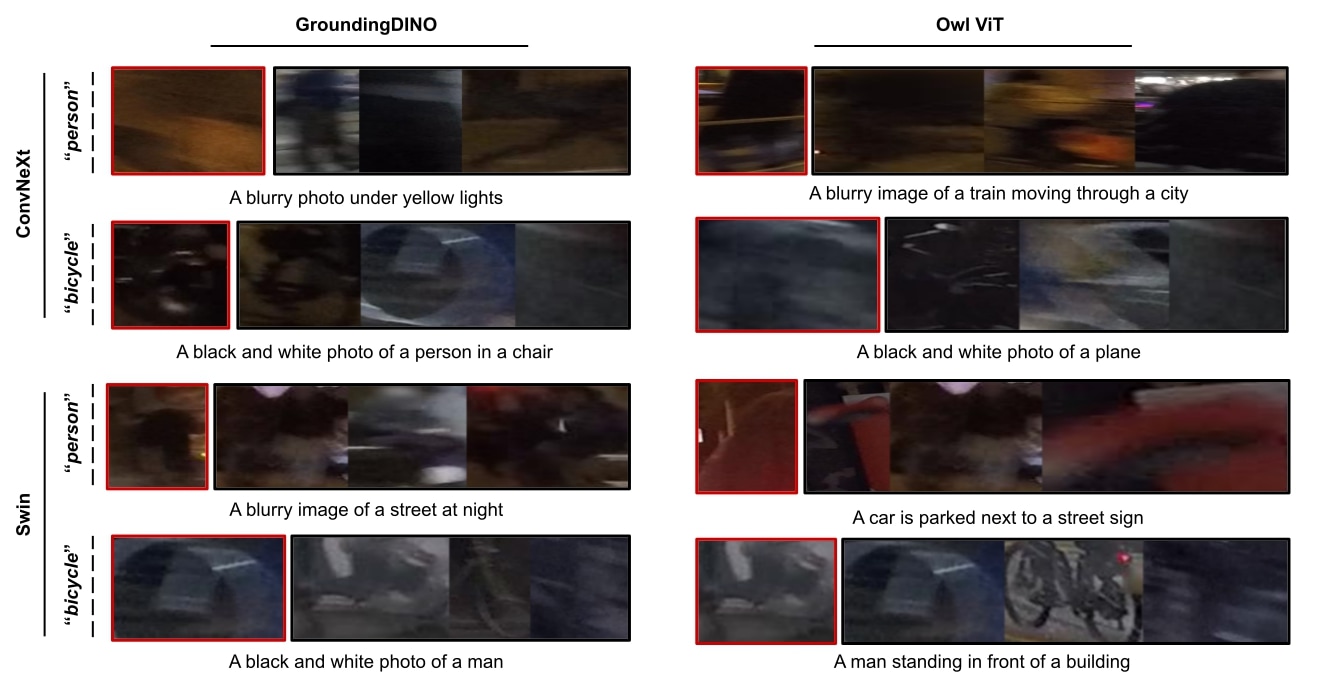}}
    \caption{
    Qualitative results of systematic error discovery in ACDC. 
    %
    %
    Our framework predicts very few false positives and discerns true negatives accurately, showing that its robustness in detecting interpretable systematic errors.
    %
    %
    \vspace{-10pt}
    }
    \label{fig:acdc_qual}
\end{figure*}
\paragraph{Results:} We show the results of discovering systematic errors for the ``\textit{person}''and ``\textit{bicycle}'' classes across all models and datasets in Table~\ref{tab:sys_acc}. Additionally, Fig.~\ref{fig:bdd_person_qual} and Fig.~\ref{fig:acdc_qual}
%
%
qualitatively present systematic errors discovered in the BDD and ACDC datasets respectively, along with language level explanations given by the captions of the query patches. 
%
%
%
For the ConvNext model evaluated on the BDD dataset (in-distribution), we see more than 95.23\% 
%
and 80.72\% 
%
%
accuracy for predicting the systematic errors for the semantic class of ``\textit{person}'' and ``\textit{bicycle}'' respectively. Similarly for the Swin SSM, accuracies corresponding to these semantic classes also relatively similar at 89.55\% 
%
%
and 69\%. 
%
%
As can be seen, both the SSMs misclassify multiple patches of \textit{snow} as ``\textit{person}''. This occurs for the \textit{car} object as well. Certain low-level dissimilarities between the caption of the query patch and specific nearest neighbor patches can also be seen. However, the query patch and its neighborhood exhibit the core systematic concept of \textit{car}.
Similarly, we find \textit{metal} objects and \textit{car} parts are systematically misclassified as ``\textit{bicycle}''. The concept of \textit{reflection} is also interpreted from the errors; however, BLIP-2 associates the origins of these reflections with water while generating the caption. We find that during human evaluation, these reflections are instead on bonnets of vehicles, which link back to the \textit{metal} and \textit{car} concept. In all instances, the interpretation of the identified systematic errors shows an above-average level of human satisfaction.
In ACDC, patches do not contain systematic errors (verified by our human evaluation). In fact, the patches only contain random errors that are not human-coherent. This is reflected in the our framework's qualitative assessment of ACDC given in Fig~\ref{fig:acdc_qual}.
Further, our algorithm shows an accuracy of 84.21\%  and 81.25\%  for ConvNeXt and 93.54\% and 58.33\% for Swin respectively on ACDC (for the ``\textit{person}'' and ``\textit{bicycle} classes). In all of these predictions, we find that the model predicts either 0 or a tiny number of systematic errors (false positives),
%
%
which shows that the algorithm effectively detects systematic errors when present and rejects them whenever not present. Our framework thus retrieves and interprets the systematic errors, and rejects non-systematic errors with high accuracy. Further, it does so even when the minimum patch size is varied from its default value of $60\times 60$. Similarly, it also performs well when $q$ is varied from its default value of $3$. These additional results are given in the supplement.
%
\section{Conclusion}
We investigate an underexplored avenue by presenting a novel framework to identify and explain interpretable systematic errors in semantic segmentation. Notably, our framework is highly modular and ablation-ready, as demonstrated by our experiments with multiple models for each intermediate step. Our proposed algorithm shows accurately localized highly similar concepts.
These factors shown the applicability of our framework as a necessary precursor to SSM deployment at industrial levels. Moreover, with improvements in SSM performance nearly saturating on existing protocols, it becomes important to assess their weaknesses using methods which are automatic and which go beyond performance-based measures. We believe that our work is an important step towards filling this gap and hope to inspire future work.
%

%
\label{discussion_conclusion}
\bibliography{aaai25}
\clearpage
\appendix
\noindent\textbf{\huge{Supplementary Material}}\\

%
%
\noindent This supplementary material contains additional results and implementation details. We mention them in the order that they are presented in the paper. First, Sec.~\ref{eval} describes our evaluation schema for identifying precision errors and systematic errors. Further, we provide results for various ablations studies in  Sec.~\ref{results} which are referred to at the end of the main manuscript. 
\section{Evaluation}
\label{eval}
\subsection{Precision error evaluation} 
We evaluate the precision errors identified by GroundingDINO by using the intersection over union metric (IoU) with the ground truth class maps.
%
%
We provide an algorithm form of our evaluation method separately for evaluating positives and negatives classified by GroundingDINO in Alg.~\ref{pos_algo}  and Alg.~\ref{neg_algo} respectively.
\begin{algorithm}[h!]
    \caption{Procedure for evaluating positives.}
    \label{pos_algo}
\begin{algorithmic}[1]
    \Procedure{EvaluatePositive}{$p$}
    \State gt $\leftarrow$ \Call{GetGroundTruthFor}{$p$}
    \State bbox $\leftarrow$ \Call{GetLocationFor}{$p$}
    \State gtpatch $\leftarrow$ \Call{GoToLocation}{gt, bbox}
    \State patchmap $\leftarrow$ \Call{MakeBinaryForClass}{$c_j$, $p$}
    \State gtmap $\leftarrow$ \Call{MakeBinaryForClass}{$c_j$, gtpatch}
    \State iou $\leftarrow$ \Call{IoU}{patchmap, gtmap}
    \If{iou $> 0.7$}
        \State \textbf{return} FP \Comment{false positive}
    \Else
        \State \textbf{return} TP \Comment{true positive}
    \EndIf
    \EndProcedure
\end{algorithmic}
\end{algorithm}
\begin{algorithm}[h!]
    \caption{Procedure for evaluating negatives.}
    \label{neg_algo}
\begin{algorithmic}[1]
    \Procedure{EvaluateNegative}{$p$}
    \State gt $\leftarrow$ \Call{GetGroundTruthFor}{$p$}
    \State bbox $\leftarrow$ \Call{GetLocationFor}{$p$}
    \State gtpatch $\leftarrow$ \Call{GoToLocation}{gt, bbox}
    \State patchmap $\leftarrow$ \Call{MakeBinaryForClass}{$c_j$, $p$}
    \State gtmap $\leftarrow$ \Call{MakeBinaryForClass}{$c_j$, gtpatch}
    \State iou $\leftarrow$ \Call{IoU}{patchmap, gtmap}
    \If{iou $> 0.7$}
        \State \textbf{return} TN \Comment{true negative}
    \Else
        \State \textbf{return} FN \Comment{false negative}
    \EndIf
    \EndProcedure
\end{algorithmic}
\end{algorithm}
%
%
%
\\
\subsection{Systematic error evaluation}
We conduct a human study in order to evaluate systematic errors identified by our algorithm. For this task, $2$ human evaluators with normal/correct vision were used to classify each precision error patch image into two classes: \textit{systematic error} (the positive class) and \textit{non-systematic error} (negative class). This evaluation was done on a $1920 \times 1080$p, $75$ Hz Samsung UHD monitor.   
\section{Ablation Studies}
\label{results}
\subsection{Varying minimum patch size in precision error identification}
\begin{figure*}
        \centering
        \includegraphics[width=\linewidth]{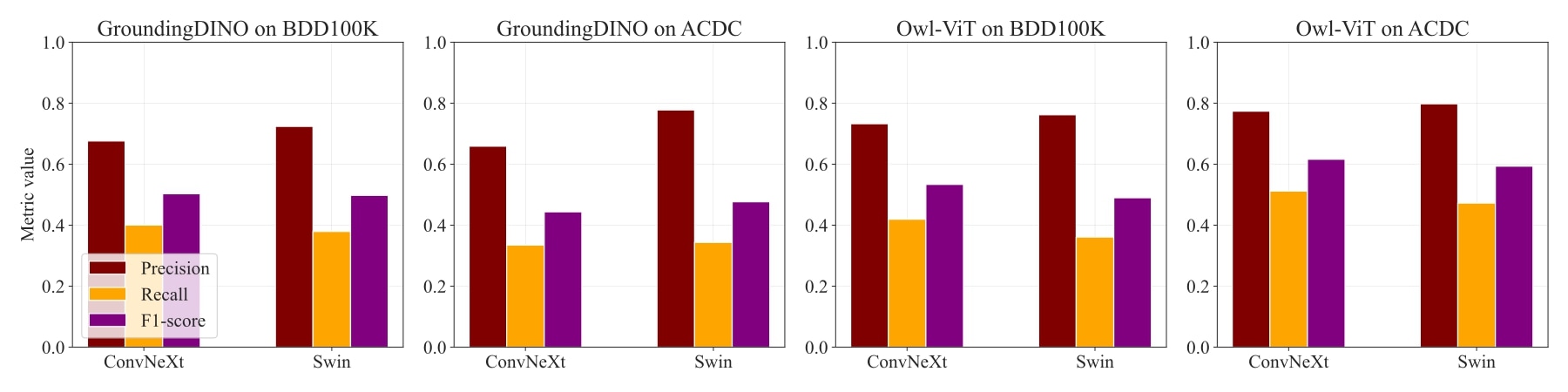}
    \caption{Precision, recall, and F1-score metrics for precision error identification for ``\textit{person}'' where $a = 40$.}
    \label{fig:person_metrics_40}
\end{figure*}

\begin{figure*}
        \centering
        \includegraphics[width=\linewidth]{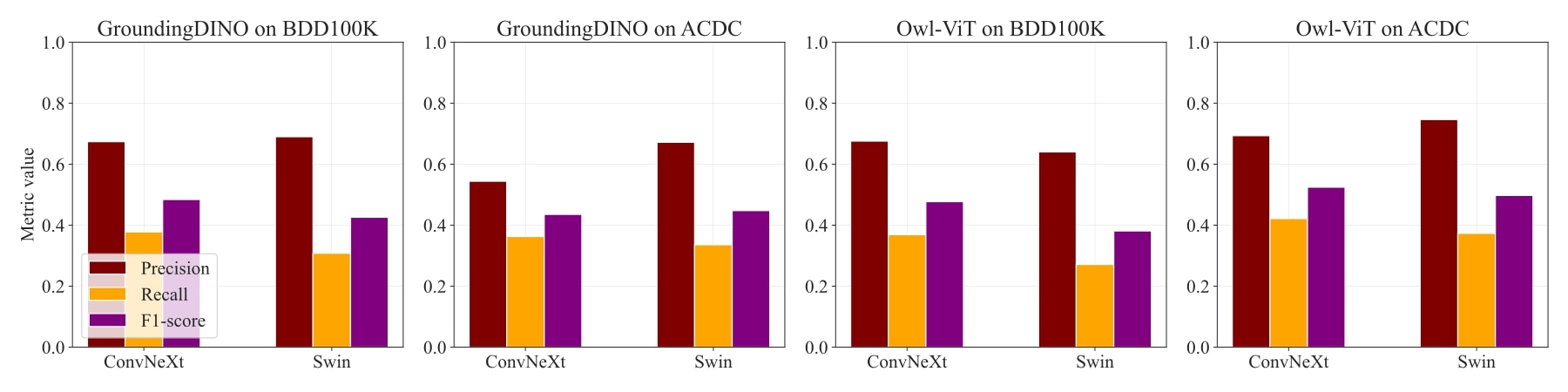}
    \caption{Precision, recall, and F1-score metrics for precision error identification for ``\textit{person}'' where $a = 60$.}
    \label{fig:person_metrics_60}
\end{figure*}

\begin{figure*}
        \centering
        \includegraphics[width=\linewidth]{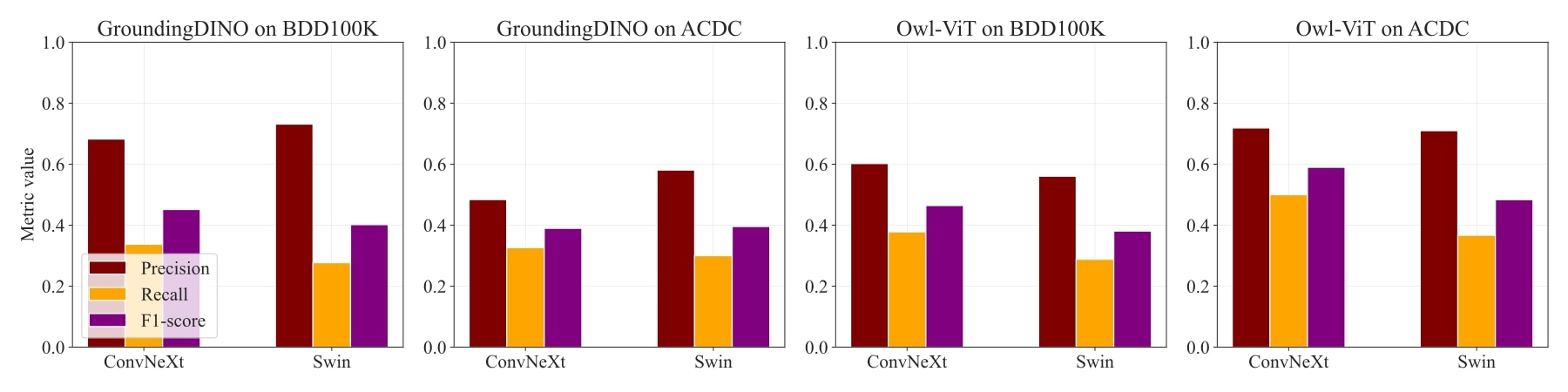}
    \caption{Precision, recall, and F1-score metrics for precision error identification for ``\textit{person}'' where $a = 80$.}
    \label{fig:person_metrics_80}
\end{figure*}

\begin{figure*}
        \centering
        \includegraphics[width=\linewidth]{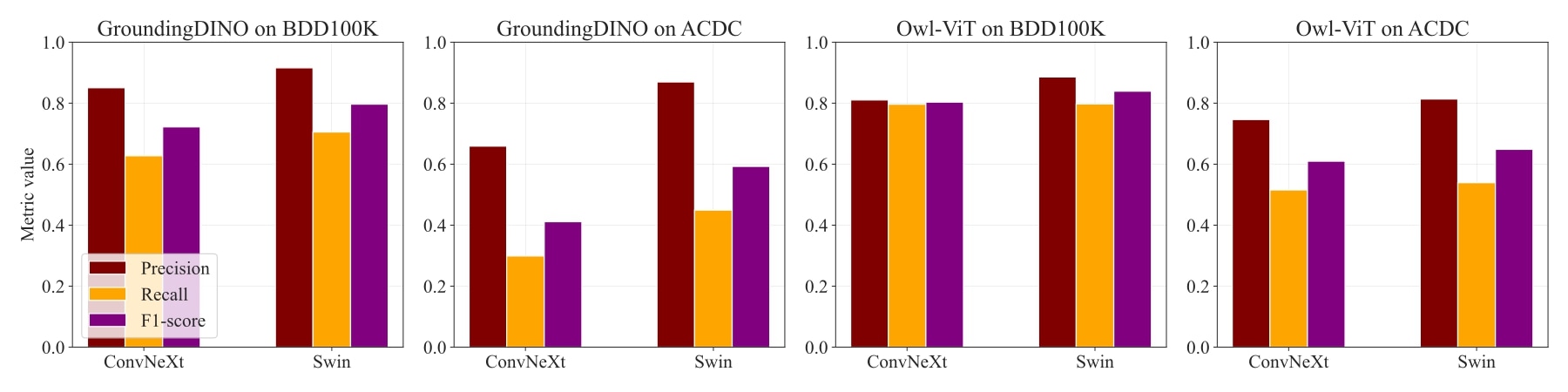}
    \caption{Precision, recall, and F1-score metrics for precision error identification for ``\textit{bicycle}'' where $a = 40$.}
    \label{fig:bicycle_metrics_40}
\end{figure*}

\begin{figure*}
        \centering
        \includegraphics[width=\linewidth]{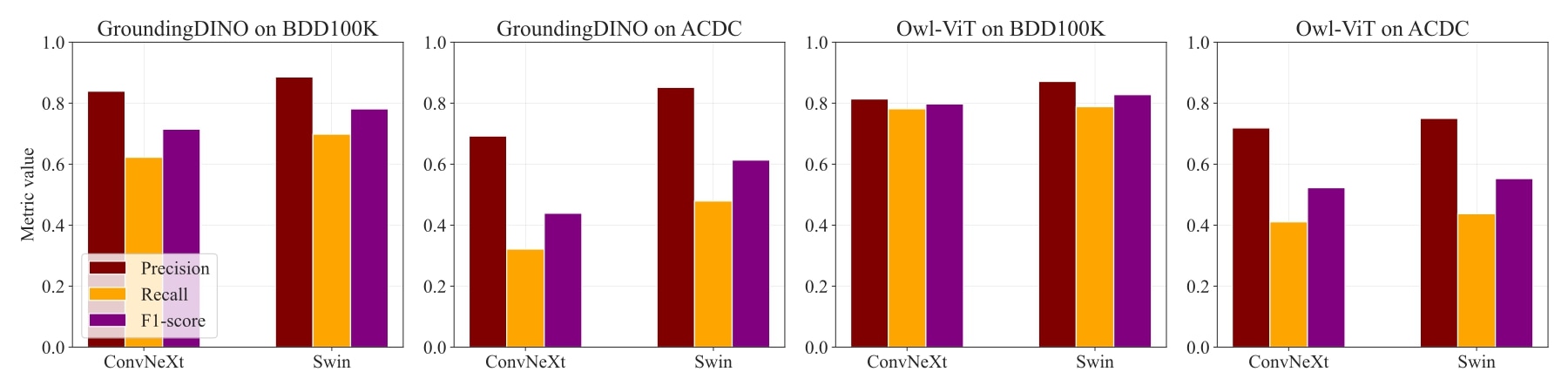}
    \caption{Precision, recall, and F1-score metrics for precision error identification for ``\textit{bicycle}'' where $a = 60$.}
    \label{fig:bicycle_metrics_60}
\end{figure*}

\begin{figure*}
        \centering
        \includegraphics[width=\linewidth]{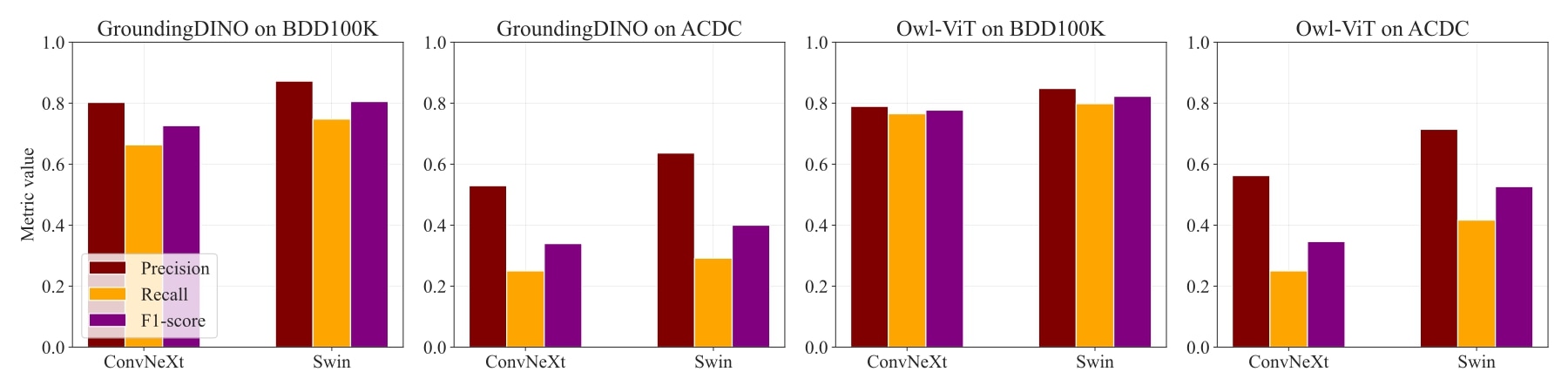}
    \caption{Precision, recall, and F1-score metrics for precision error identification for ``\textit{bicycle}'' where $a = 80$.}
    \label{fig:bicycle_metrics_80}
\end{figure*}

Following Sec.~\textcolor{red}{4.2}, our framework utilizes a multimodal object detection model $G(\cdot, \cdot)$ to infer the presence of the semantic class $c_j$ in a given patch $p$. 
This patch is constrained to have a minimum image size of $a \times a$, which is varied to $40 \times 40, 60 \times 60, 80 \times 80$ as an ablation study.
We evaluate each detector's (GroundingDINO and Owl ViT) ability to identify errors for all values of $a \in \{40, 60, 80\}$ in terms of accuracy in Table~\ref{tab:detector_bench_40}, Table~\ref{tab:detector_bench_60}, Table~\ref{tab:detector_bench_80} for both ``\textit{person}'' and ``\textit{bicycle}'' classes. 
Additionally, precision, recall and f1-score metrics for these evaluations are presented in Fig.~\ref{fig:person_metrics_40}, Fig.~\ref{fig:person_metrics_60}, Fig.~\ref{fig:person_metrics_80} for all values of $a$ where $c_j=$``\textit{person}``. 
Similarly, Fig.~\ref{fig:bicycle_metrics_20}, Fig.~\ref{fig:bicycle_metrics_40},Fig.~\ref{fig:bicycle_metrics_60}, and Fig.~\ref{fig:bicycle_metrics_80} report these metrics for all values of $a$ where $c_j$=``\textit{bicycle}''.\\ 

\begin{table}[H]
    \centering
    \resizebox{\linewidth}{!}{
    \begin{tabular}{cccccc}
        \toprule

        \multirow{2}{*}{Test data} & \multirow{2}{*}{Semantic class} & \multicolumn{2}{c}{GroundingDINO} & \multicolumn{2}{c}{Owl ViT}\\
        \cmidrule{3-4} \cmidrule{5-6}
         & & ConvNeXt & Swin & ConNeXt & Swin\\
         \midrule
        \multirow{2}{*}{BDD100k \tiny{(in-distribution)}} & person & 64.61 & 62.23 & 67.20 & 62.90\\
         & bicycle & 64.38 & 70.77 & 71.25 & 75.16\\
        \multirow{2}{*}{ACDC \tiny{(out-distribution)}} & person & 57.23 & 56.23 & 67.47 & 62.46\\
         & bicycle & 43.92 & 59.56 & 56.76 & 61.76\\
        \bottomrule
    \end{tabular}
    }
    \caption{Benchmarking GroundingDINO and Owl-ViT on precision error identification with minimum patch size $= 40$. All reported values follow the accuracy metric.}
    \label{tab:detector_bench_40}
\end{table}
\begin{table}[H]
    \centering
    \resizebox{\linewidth}{!}{
    \begin{tabular}{cccccc}
        \toprule

        \multirow{2}{*}{Test data} & \multirow{2}{*}{Semantic class} & \multicolumn{2}{c}{GroundingDINO} & \multicolumn{2}{c}{Owl ViT}\\
        \cmidrule{3-4} \cmidrule{5-6}
         & & ConvNeXt & Swin & ConNeXt & Swin\\
         \midrule
        \multirow{2}{*}{BDD100k \tiny{(in-distribution)}} & person & 69.77 & 66.85 & 69.66 & 64.83\\
         & bicycle & 65.44 & 70.28 & 72.35 & 75.10\\
        \multirow{2}{*}{ACDC \tiny{(out-distribution)}} & person & 61.75 & 59.93 & 68.92 & 63.54\\
         & bicycle & 49.45 & 65.88 & 53.85 & 60.00\\
        \bottomrule
    \end{tabular}
    }
    \caption{Benchmarking GroundingDINO and Owl-ViT on precision error identification with minimum patch size $= 60$. All reported values follow the accuracy metric.}
    \label{tab:detector_bench_60}
\end{table}
\begin{table}[H]
    \centering
    \resizebox{\linewidth}{!}{
    \begin{tabular}{cccccc}
        \toprule

        \multirow{2}{*}{Test data} & \multirow{2}{*}{Semantic class} & \multicolumn{2}{c}{GroundingDINO} & \multicolumn{2}{c}{Owl ViT}\\
        \cmidrule{3-4} \cmidrule{5-6}
         & & ConvNeXt & Swin & ConNeXt & Swin\\
         \midrule
        \multirow{2}{*}{BDD100k \tiny{(in-distribution)}} & person & 74.32 & 72.24 & 72.68 & 68.44\\
         & bicycle & 67.33 & 74.25 & 71.73 & 75.45\\
        \multirow{2}{*}{ACDC \tiny{(out-distribution)}} & person & 66.19 & 63.82 & 76.98 & 69.08\\
         & bicycle & 46.97 & 59.62 & 48.48 & 65.38\\
        \bottomrule
    \end{tabular}
    }
    \caption{Benchmarking GroundingDINO and Owl-ViT on precision error identification with minimum patch size $= 80$. All reported values follow the accuracy metric.}
    \label{tab:detector_bench_80}
\end{table}
\subsection{Varying minimum patch size in systematic error assessment}
Similarly, we vary the minimum patch size in the final stage of our algorithm, \textit{i.e.}, systematic error detection. The minimum size of the patches, $a \times a$ is varied to $40 \times 40$, $60 \times 60$, and $80 \times 80$. Following this, the systematic error assessment experiment is repeated, the results of which are presented in Table~\ref{tab:sys_eval_40}, Table~\ref{tab:sys_eval_60}, and Table~\ref{tab:sys_eval_80} below.
\begin{table}[H]
    \centering
    \resizebox{\linewidth}{!}{
    \begin{tabular}{cccccc}
        \toprule
        \multirow{2}{*}{Test data} & \multirow{2}{*}{Semantic class} & \multicolumn{2}{c}{GroundingDINO} & \multicolumn{2}{c}{Owl ViT}\\
        \cmidrule{3-4} \cmidrule{5-6}
         & & ConvNeXt & Swin & ConNeXt & Swin\\
         \midrule
        \multirow{2}{*}{BDD100k \tiny{(in-distribution)}} & person & 90.00 & 91.95 & 88.54 & 81.97\\
         & bicycle & 69.16 & 55.79 & 73.33 & 67.27\\
        \multirow{2}{*}{ACDC \tiny{(out-distribution)}} & person & 82.17 & 88.89 & 80.90 & 80.83\\
         & bicycle & 75.00 & 69.56 & 74.63 & 66.10\\
        \bottomrule
    \end{tabular}
    }
    \caption{Benchmarking GroundingDINO and Owl-ViT on systematic error assessment with minimum patch size $= 40$. All reported values follow the accuracy metric.}
    \label{tab:sys_eval_40}
\end{table}
\begin{table}[H]
    \centering
    \resizebox{\linewidth}{!}{
    \begin{tabular}{cccccc}
        \toprule
        \multirow{2}{*}{Test data} & \multirow{2}{*}{Semantic class} & \multicolumn{2}{c}{{GroundingDINO}} & \multicolumn{2}{c}{{Owl ViT}}\\
        \cmidrule{3-4} \cmidrule{5-6}
         & & {{ConvNeXt}} & {{Swin}} & {{ConNeXt}} & {{Swin}}\\
         \midrule
        \multirow{2}{*}{{BDD100k \tiny{(in-distribution)}}} & {{person}} & 95.23 & 89.55 & 84.86 & 82.67\\
         & {\small{bicycle}} & 80.72 & 69.00 & 74.48 & 67.84\\
        \multirow{2}{*}{{ACDC \tiny{(out-of-distribution)}}} & {{person}} & 84.21 & 93.54 & 83.87 & 88.06\\
         & {\small{bicycle}} & 81.25 & 58.33 & 75.00 & 57.14\\
        \bottomrule
    \end{tabular}
    }
    \caption{Benchmarking GroundingDINO and Owl-ViT on systematic error assessment with minimum patch size $= 60$. All reported values follow the accuracy metric.}
    \label{tab:sys_eval_60}
\end{table}
\begin{table}[H]
    \centering
    \resizebox{\linewidth}{!}{
    \begin{tabular}{cccccc}
        \toprule
        \multirow{2}{*}{Test data} & \multirow{2}{*}{Semantic class} & \multicolumn{2}{c}{GroundingDINO} & \multicolumn{2}{c}{Owl ViT}\\
        \cmidrule{3-4} \cmidrule{5-6}
         & & ConvNeXt & Swin & ConNeXt & Swin\\
         \midrule
        \multirow{2}{*}{BDD100k \tiny{(in-distribution)}} & person & 95.83 & 89.41 & 91.19 & 84.26\\
         & bicycle & 65.43 & 61.76 & 66.32 & 60.71\\
        \multirow{2}{*}{ACDC \tiny{(out-distribution)}} & person & 90.32 & 93.55 & 81.25 & 83.87\\
         & bicycle & 82.35 & 54.54 & 81.25 & 64.29\\
        \bottomrule
    \end{tabular}
    }
    \caption{Benchmarking GroundingDINO and Owl-ViT on systematic error assessment with minimum patch size $= 80$. All reported values follow the accuracy metric.}
    \label{tab:sys_eval_80}
\end{table}
\subsection{Varying the number of nearest neighbours in systematic error assessment}
Additionally, we also vary the value of $q$, \textit{i.e.}, the number of nearest neighbours of the query patch during our systematic error analysis such that $q \in \{3,5,7\}$. This is presented in a manner similar to the main paper in Table~\ref{tab:sys_eval_3}, Table~\ref{tab:sys_eval_5}, and Table~\ref{tab:sys_eval_5} below. Note that the minimum patch here is set to the default value of $60 \times 60$. We see that with values of $q$ higher than $3$, the ``\textit{bicycle}'' class for the BDD dataset suffers particularly. This is because un-interpretable patches also make their way into the nearest neighbours of coherent patches and reduce the conceptual quality of the query patch and its nearest neighbour family.
\begin{table}[H]
    \centering
    \resizebox{\linewidth}{!}{
    \begin{tabular}{cccccc}
        \toprule
        \multirow{2}{*}{Test data} & \multirow{2}{*}{Semantic class} & \multicolumn{2}{c}{{GroundingDINO}} & \multicolumn{2}{c}{{Owl ViT}}\\
        \cmidrule{3-4} \cmidrule{5-6}
         & & {{ConvNeXt}} & {{Swin}} & {{ConNeXt}} & {{Swin}}\\
         \midrule
        \multirow{2}{*}{{BDD100k \tiny{(in-distribution)}}} & {{person}} & 95.23 & 89.55 & 84.86 & 82.67\\
         & {\small{bicycle}} & 80.72 & 69.00 & 74.48 & 67.84\\
        \multirow{2}{*}{{ACDC \tiny{(out-of-distribution)}}} & {{person}} & 84.21 & 93.54 & 83.87 & 88.06\\
         & {\small{bicycle}} & 81.25 & 58.33 & 75.00 & 57.14\\
        \bottomrule
    \end{tabular}
    }
    \caption{Benchmarking GroundingDINO and Owl-ViT on systematic error assessment with $q=3$. All reported values follow the accuracy metric.}
    \label{tab:sys_eval_3}
\end{table}
\begin{table}[H]
    \centering
    \resizebox{\linewidth}{!}{
    \begin{tabular}{cccccc}
        \toprule
        \multirow{2}{*}{Test data} & \multirow{2}{*}{Semantic class} & \multicolumn{2}{c}{{GroundingDINO}} & \multicolumn{2}{c}{{Owl ViT}}\\
        \cmidrule{3-4} \cmidrule{5-6}
         & & {{ConvNeXt}} & {{Swin}} & {{ConNeXt}} & {{Swin}}\\
         \midrule
        \multirow{2}{*}{{BDD100k \tiny{(in-distribution)}}} & {{person}} & 88.42 & 86.08 & 84.86 & 80.56\\
         & {\small{bicycle}} & 56.25 & 61.75 & 62.76 & 66.08\\
        \multirow{2}{*}{{ACDC \tiny{(out-of-distribution)}}} & {{person}} & 91.17 & 91.04 & 83.87 & 86.56\\
         & {\small{bicycle}} & 80.77 & 59.26 & 81.25 & 53.57 \\
        \bottomrule
    \end{tabular}
    }
    \caption{Benchmarking GroundingDINO and Owl-ViT on systematic error assessment with $q=5$. All reported values follow the accuracy metric.}
    \label{tab:sys_eval_5}
\end{table}
\begin{table}[H]
    \centering
    \resizebox{\linewidth}{!}{
    \begin{tabular}{cccccc}
        \toprule
        \multirow{2}{*}{Test data} & \multirow{2}{*}{Semantic class} & \multicolumn{2}{c}{GroundingDINO} & \multicolumn{2}{c}{Owl ViT}\\
        \cmidrule{3-4} \cmidrule{5-6}
         & & ConvNeXt & Swin & ConNeXt & Swin\\
         \midrule
        \multirow{2}{*}{BDD100k \tiny{(in-distribution)}} & person & 89.47 & 85.44 & 83.24 & 71.52\\
         & bicycle & 57.14 & 55.03 & 58.62 & 54.49\\
        \multirow{2}{*}{ACDC \tiny{(out-distribution)}} & person & 89.71 & 89.55 & 87.09 & 86.57\\
         & bicycle & 80.76 & 59.26 & 84.37 & 53.57\\
        \bottomrule
    \end{tabular}
    }
    \caption{Benchmarking GroundingDINO and Owl-ViT on systematic error assessment with $q=7$. All reported values follow the accuracy metric.}
    \label{tab:sys_eval_7}
\end{table}
%
%
%
%
%
%
\setlength{\arrayrulewidth}{5pt}
\setlength{\arrayrulewidth}{5pt}
\setlength{\arrayrulewidth}{5pt}
%
%
%
%
%
\setlength{\arrayrulewidth}{5pt}
\setlength{\arrayrulewidth}{5pt}
\setlength{\arrayrulewidth}{5pt}
\end{document}